\pdfoutput=1
\documentclass[10pt,twocolumn,letterpaper]{article}

\usepackage{iccv}
\usepackage{times}
\usepackage{epsfig}
\usepackage{graphicx}
\usepackage{amsmath}
\usepackage{amssymb}
\graphicspath{{figures/}}

\usepackage[pagebackref=true,breaklinks=true,letterpaper=true,colorlinks,bookmarks=false]{hyperref}

\iccvfinalcopy 

\def\iccvPaperID{12758} 

\ificcvfinal\fi

\begin{document}

\title{Contextual Affinity Distillation for Image Anomaly Detection}

\author{Jie Zhang$^1$
~~~~Masanori Suganuma$^1$
~~~~Takayuki Okatani$^{1,2}$\\
$^1$Graduate School of Information Sciences, Tohoku University ~~~~ $^2$RIKEN Center for AIP \\
{\tt\small \{jzhang,suganuma,okatani\}@vision.is.tohoku.ac.jp}
}

\maketitle
\ificcvfinal\thispagestyle{empty}\fi

\begin{abstract}
Previous works on unsupervised industrial anomaly detection mainly focus on local structural anomalies such as cracks and color contamination. While achieving significantly high detection performance on this kind of anomaly, they are faced with logical anomalies that violate the long-range dependencies such as a normal object placed in the wrong position. In this paper, based on previous knowledge distillation works, we propose to use two students (local and global) to better mimic the teacher's behavior. The local student, which is used in previous studies mainly focuses on structural anomaly detection while the global student pays attention to logical anomalies. To further encourage the global student's learning to capture long-range dependencies, we design the global context condensing block (GCCB) and propose a contextual affinity loss for the student training and anomaly scoring. Experimental results show the proposed method doesn't need cumbersome training techniques and achieves a new state-of-the-art performance on the MVTec LOCO AD dataset.
\end{abstract}

\section{Introduction}
\label{sec:intro}
The task of anomaly detection and localization aims to identify whether an image is normal or anomalous and localize the anomalies \cite{chalapathy2019deep,ruff2021unifying}. It has a wide range of real-world applications including industrial inspection of products \cite{bergmann2019mvtec,bergmann2018improving}. As anomalous samples rarely appear in manufacturing product lines and the unpredictable nature of anomalies, most of the efforts are paid to unsupervised anomaly detection methods, in which we have only anomaly-free samples for training.

Recent studies showed that using intermediate features of a deep pre-trained model is representative enough to achieve state-of-the-art performance \cite{reiss2021panda}. Knowledge distillation \cite{hinton2015distilling} is a straightforward and effective way to achieve this goal. Recent knowledge distillation-based anomaly detection approaches \cite{bergmann2020uninformed,wang2021student,salehi2021multiresolution,deng2022anomaly,bergmann2022beyond} try to transfer the knowledge of normal samples from a teacher which is pre-trained on a large-scale natural image dataset, \textit{e}.\textit{g}. ImageNet \cite{deng2009imagenet} into a student model. The use of per-pixel \cite{wang2021student,deng2022anomaly} or local patch-based regression loss \cite{bergmann2020uninformed,bergmann2022beyond} further improves the fine-grained knowledge transfer and anomaly detection performance. The teacher model acts as a knowledgeable feature extractor that could extract representative feature embeddings for both normal and anomalous samples, while the student is trained exclusively on anomaly-free samples that it is expected to only mimic the teacher's behavior for normal features. During inference, the anomaly scores are derived from the discrepancy between student and teacher features. As anomalies could be of any size and at any abstract level, using multi-layer features from the teacher could better cover more types of anomalies.

\begin{figure}[t]
  \centering
   \includegraphics[width=0.95\linewidth]{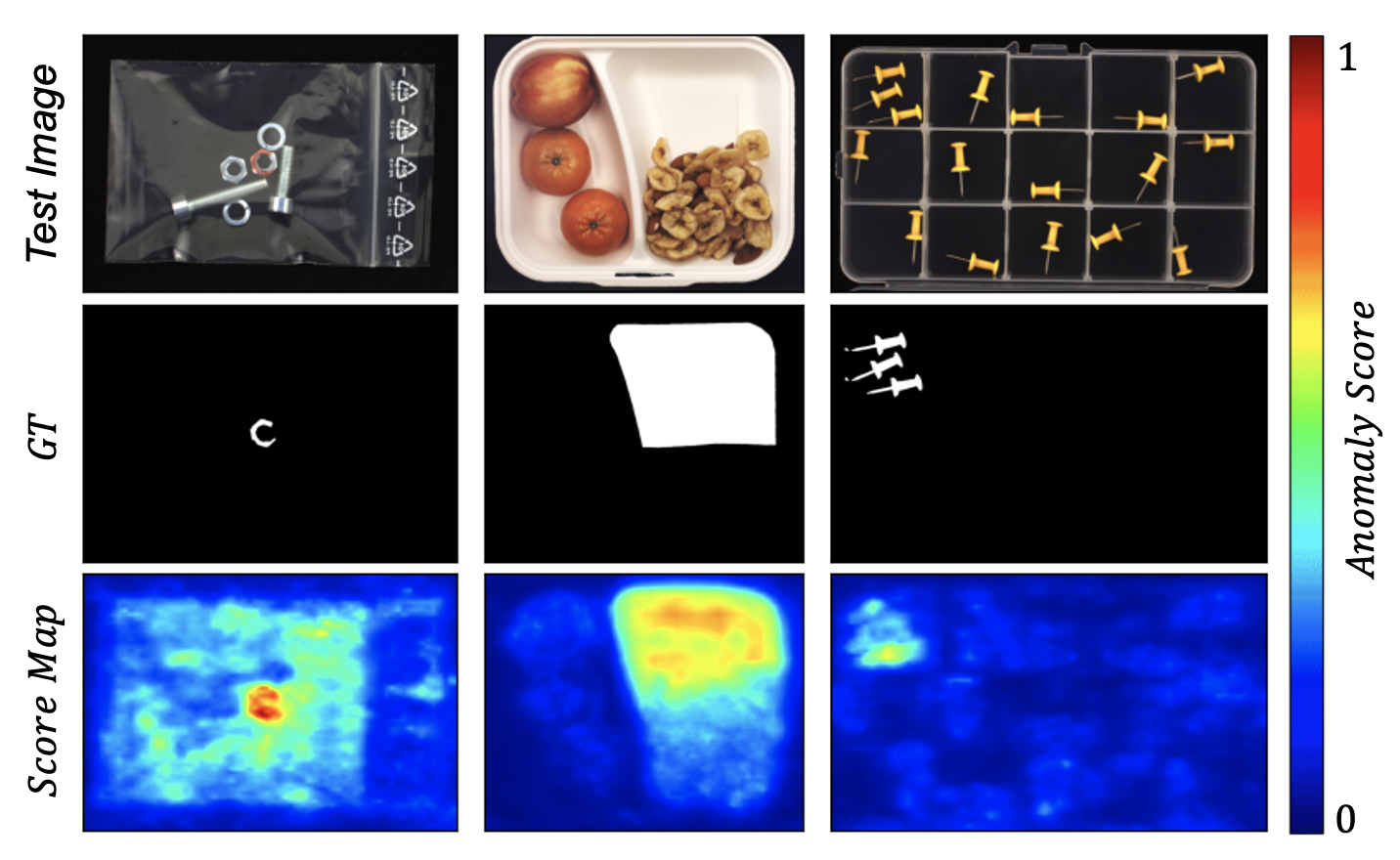}

   \caption{Examples from MVTec LOCO \cite{bergmann2022beyond}. We show structural and logical anomaly samples with our detection results. Our method could detect both kinds of anomalies.}
   \label{fig_examples}
\end{figure}

Previous unsupervised anomaly detection and localization datasets focus on concise scenes where each image consists of only one product object, \textit{e}.\textit{g}. a capsule or one kind of texture. In this case, most of the anomalies are local \textit{structural} anomalies such as cracks and scratches. The above-mentioned methods are effective for this structural anomaly detection. However, in complex scenarios with global contextual constraints, \textit{logical} anomalies violating long-range dependencies such as a normal object appearing in the wrong position or a missing object are likely to be identified as normal. Figure \ref{fig_examples} shows structural and logical anomaly examples from MVTec LOCO \cite{bergmann2022beyond} dataset. The first image from the screw bag category contains contamination as a structural anomaly, while cereals missing in the breakfast box and two additional pushpins in one compartment are defined as logical anomalies. However, directly using deep high-semantic level features from a pre-trained model cannot address this issue well, as deeper features are more source domain biased \cite{roth2022towards} and could generalize to anomaly features \cite{deng2022anomaly}.

To better detect both structural and logical anomalies, we propose to further use a global student in reverse distillation method \cite{deng2022anomaly} to comprise the dual-student knowledge distillation framework (DSKD). Unlike conventional ensemble methods \cite{dietterich2000ensemble,rokach2010ensemble,lakshminarayanan2017simple} in which each model has the same role as each other, we explicitly divide the student models into two models: {\it local} and {\it global} students. The objective of the local student is to detect structural anomalies, and it is trained to reconstruct the low-level features of those of the teacher model. The global student is trained to capture global contextual information for detecting logical anomalies. Since the logical anomalies are harder to detect than the structural anomalies, we introduce a global context condensing block into the global student, aiming to capture global information from images effectively \cite{liu2020towards,bergmann2022beyond}. The training paradigm of our method is based on the reverse distillation \cite{deng2022anomaly} where the teacher acts as an encoder and the students play the role of decoders for the feature reconstruction.

Moreover, we propose a contextual affinity loss to promote further capturing the global contextual information by the global student. Specifically, we compute the cosine similarity between each feature vector of the global student and all feature vectors of the teacher, and then the cosine similarity maps are converted into the probability distribution by a softmax function. We minimize the discrepancy between the probability distributions of the global student and the teacher. It should be noted that this differs from simply choosing several neighboring feature representations such as pair-wise distillation \cite{liu2019structured} in that such a pair-wise loss treats all features equally, but ours can capture important contextual information at the whole image level.

We conduct extensive experiments on public unsupervised anomaly detection and localization datasets and achieve state-of-the-art performance. Our contributions are threefold:
\begin{enumerate}
    \item We introduce a new dual-student knowledge distillation framework for anomaly detection. The local student aims for accurate local feature reconstruction and the global student focuses on global contextual information. The students play different roles, enabling the detection ability for both structural and logical anomalies.
    \item We propose the global contextual condensing block and contextual affinity loss, further enforcing the global contextual learning ability.
    \item We demonstrate the effectiveness of our proposed method and report new state-of-the-art performance.
\end{enumerate}



\begin{figure*}
  \centering
    \includegraphics[width=0.9\linewidth]{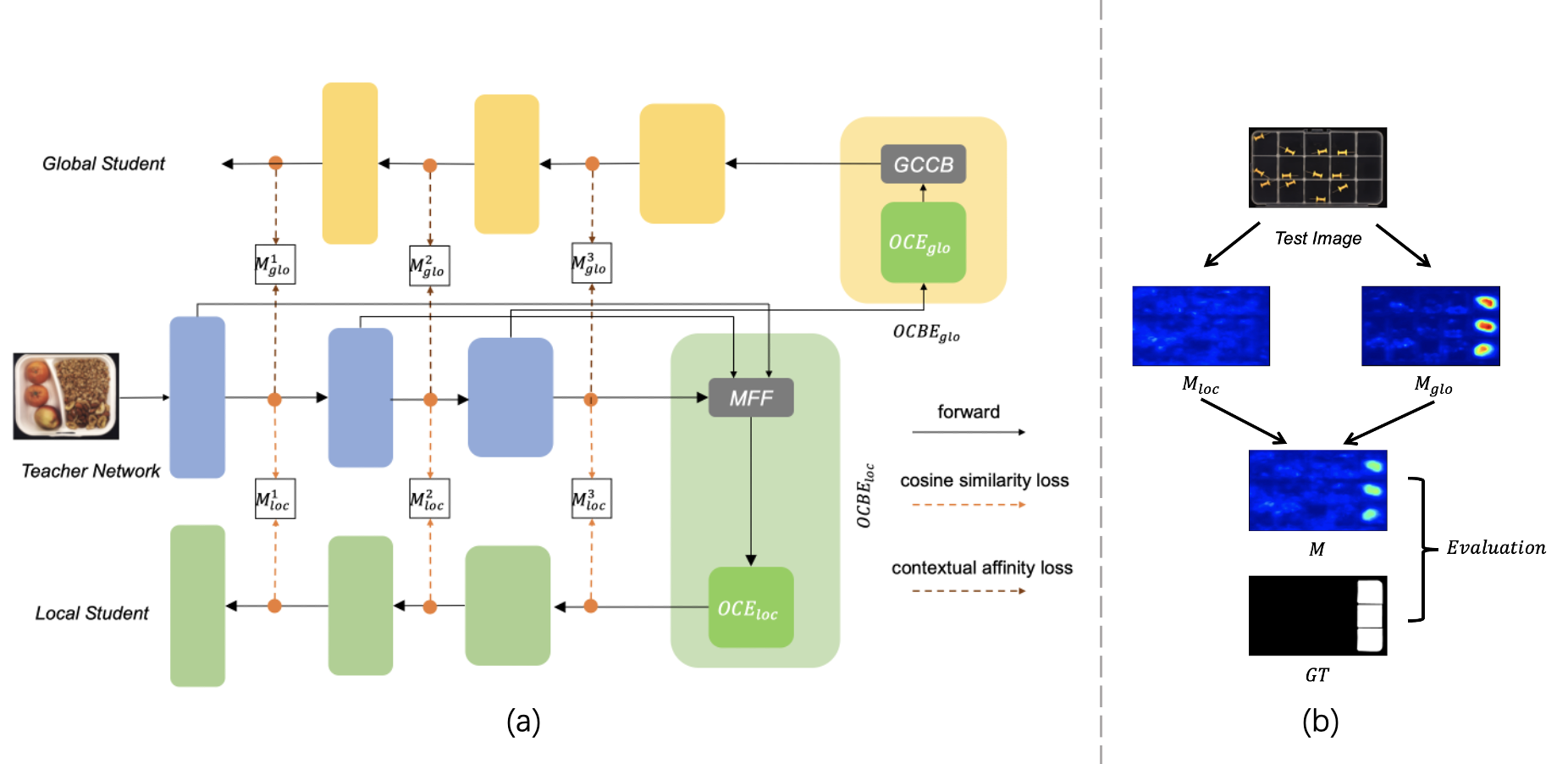}

  \caption{Overview of the dual-student knowledge distillation framework. (a) Our model employs a pre-trained teacher encoder as the feature extractor $T$, a local student for accurate low-level feature learning, and a global student to capture global contextual information. During training, the students can only learn to mimic the teacher's behavior for normal samples. (b) Anomaly scoring. Firstly, the multi-scale score maps from each student are accumulated into one single scale-normalized map separately. Then the two normalized score maps are added together to get our final detection results.}
    \label{fig_overview}
\end{figure*}

\section{Related Works}
\label{related work}
We briefly review recent research on unsupervised anomaly detection as well as related knowledge distillation works on \textit{supervised} dense prediction tasks. The recent works on 
anomaly detection could be classified into three prototypes: generative models, anomaly synthesis-based methods, and methods leveraging features extracted by pre-trained networks. 

Generative models aim to reconstruct normal samples from the encoded feature space. Autoencoders (AEs) and Gnenerative Adversarial Nets \cite{goodfellow2020generative} are popularly used for sample reconstruction. These models are trained exclusively on normal images. Since the input image is encoded into a compact feature space to only keep the most useful information, the unseen anomalies are expected to be abandoned during inference and thus reconstruct the anomaly-free images for anomaly samples \cite{bergmann2018improving}. However, deep models could generalize well to anomaly patterns and fail to detect anomalies. To overcome this issue, normal representation searching \cite{schlegl2019f} in the encoded continuous feature space, iterative reconstruction approaches \cite{dehaene2020iterative} and memory-guied autoencoders \cite{gong2019memorizing, park2020learning} are proposed to limit the model's generalization ability.

Anomaly synthesis-based methods \cite{pourreza2021g2d, li2021cutpaste, zavrtanik2021draem} focus on addressing the issue of the lack of anomaly samples so as to train the models in a supervised manner. However, the detection ability is heavily affected by the synthesizing strategies, and their performance shows a strong bias to the synthesized kind of anomalies \cite{zavrtanik2022dsr, hendrycks2018deep}. To generate more realistic anomalies, DSR \cite{zavrtanik2022dsr} tries to generate near-in-distribution low-level anomalies from a vector-quantized feature space. However, it is still challenging for high semantic-level anomaly generation.

There is also a lot of attention paid to employing pre-trained models to extract representative features from images. The kNN-based approaches \cite{cohen2020sub, roth2022towards} construct a feature gallery for normal representations and derive anomaly scores by computing the distances between input and its nearest neighbors in the feature space. They suffer from computational complexity \cite{yu2021fastflow, deng2022anomaly} and can not utilize high-semantic level features well as they as more source-domain biased \cite{roth2022towards}.

The knowledge distillation-based approaches try to transfer knowledge of normal samples to student networks. US \cite{bergmann2020uninformed} distills knowledge from a pre-trained teacher network to an ensemble of students for each patch scale. MKD \cite{salehi2021multiresolution} directly distills multi-level features into one compact student model. STFPM \cite{wang2021student} uses a vector-wise cosine similarity loss for both student training and anomaly scoring. The reverse distillation \cite{deng2022anomaly} proposed the encoder-decoder architecture to distill the knowledge from a bottleneck feature space. These methods are capable of learning local features or patches but are likely to ignore global contextual constraints. GCAD \cite{bergmann2022beyond} designed a two-branch framework based on US \cite{bergmann2020uninformed} for both structural and logical anomaly detection but it is still a two-step distillation framework where the teacher is trained with a deep pre-trained model and a large number of cropped image patches from ImageNet \cite{deng2009imagenet}. To insure training stability, multi-step training and skip connections with linearly decreased weights are added to the student.

There are also some knowledge distillation methods applied to dense prediction tasks such as semantic segmentation \cite{long2015fully, zhao2017pyramid} and object detection \cite{lin2017feature} trying to transfer the knowledge to a compact student network via fully exploring the rich information within the intermediate features from the teacher. MIMIC \cite{li2017mimicking} samples features from feature regions for object detection. Pair-wise knowledge distillation \cite{liu2019structured} was proposed to distill the structured knowledge from the feature. Channel-wise knowledge distillation \cite{shu2021channel} converts the features of each channel into probability distributions leveraging the prior that the activations from each channel tend to encode specific scene categories. It is pointed out that strictly applying the per-pixel loss which means each pixel or correlation is treated equally may enforce overly strict constraints on the student model and lead to sub-optimal solutions \cite{shu2021channel}. However, the most important guidance for training the student model comes from the ground truth labels that are not available for unsupervised anomaly detection. Also, in conventional knowledge distillation applications where only the student model is deployed after training, the structured knowledge is computed or evaluated \textit{separately} for each student and teacher feature map, while both the teacher and student are used for knowledge distillation-based anomaly detection methods. We distinguish our proposed contextual affinity loss from prior arts that the contextual affinity for student features is computed using both student and teacher features for better guidance and to make training stable.
\section{Proposed Method}
\subsection{Problem Formulation}
Given a set of anomaly-free training images $ \mathcal{S}^t=\{I_1^t, ..., I_{n_t}^t\}$ and a validation set $ \mathcal{S}^v=\{I_1^v, ..., I_{n_v}^v\}$ that consists of also anomaly-free images, we aim to detect if a test image from the test set $ \mathcal{S}^q=\{I_1^q, ..., I_{n_q}^q\}$ is anomalous or not, and also localize the defected area if it is anomalous.

\subsection{Overview of the Dual-student Framework} 
As shown in Fig \ref{fig_overview}, our dual-student distillation framework consists of five parts: a deep neural network pre-trained on ImageNet as the fixed teacher $T$ to extract multi-level representative features, a one-class bottleneck embedding module $OCBE_{loc}$ for the local student, a local student decoder $S_{loc}$, a $OCBE_{glo}$ for the global student that contains a global context condensing block $GCCB$, and a global student decoder $S_{glo}$. The $OCBE_{loc}$ is designed for fusing multi-level features into a compact feature space followed by a local student decoder $S_{loc}$ to reconstruct the feature representations, especially low-level features accurately. The first three modules compose the reverse distillation \cite{deng2022anomaly} which is effective for structural anomaly detection and we change nothing for it in order to preserve the accurate low-level feature reconstruction ability. To better capture global contextual correlations which we expect to have the benefit of logical anomaly detection, we additionally design the $S_{glo}$ with a $GCCB$ that can keep the most condensed contextual information, and the output is then decoded by the global student decoder $S_{glo}$. Since the teacher $T$ is pre-trained on a large natural image dataset, it is expected to extract representative features for both normal and anomaly inputs. However, the two students are trained solely on anomaly-free samples, and they fail to mimic the teacher's behavior for either low-level structural anomalies or global logical anomalies during inference. The pixel-level anomaly scores are computed by comparing the decoded features from both students and the features extracted by the teacher. The local student is primarily responsible for structural anomaly detection and the global student pays more attention to the global contextual constraints.
\subsection{Local Knowledge Distillation}
Different from the conventional discriminative paradigm where the student is also a feature extractor, reverse distillation \cite{deng2022anomaly} is in a generative manner to reconstruct the features extracted by the teacher. The student decoder receives dense encoded features and decodes the features from high-semantic levels to low-semantic levels. The dense feature space is likely to abandon unseen anomalous feature representations at inference to encourage feature discrepancies for anomalies. We use the reverse distillation \cite{deng2022anomaly} method for accurate feature reconstruction and structural anomaly detection.
Given an image $I$, the output of the first three residual stages of a pre-trained WideResNet50 \cite{zagoruyko2016wide} $T$ extracts multi-layer intermediate features $F_T^l \in \mathbb{R}^{h_l \times w_l \times c_l}$, where $l \in \{1, 2, 3\}$. The $OCBE_{loc}$ encodes the features into the embedding  $\phi_{loc}$. The local student $S_{loc}$ then generates the corresponding feature maps $F_{S_{loc}}^l$ from $\phi_{loc}$. The student decoder has a symmetrical architecture with the teacher $T$ while the input is high abstract feature representations and the down-sampling operations used in original ResNets \cite{he2016deep} are replaced by up-samplings. The vector-wise cosine distance is used as the loss function for training the local student. A $2-D$ anomaly score map could be obtained at each layer scale  
\begin{equation}
\label{map_loc}
    M_{loc}^l = \mathbf{1} - \frac{F_T^l \cdot F_{S_{loc}}^l}{\|F_T^l\| \| F_{S_{loc}}^l \|}
\end{equation}
The final loss for training the local student is 
\begin{equation}
    \mathcal{L}_{loc} = \sum_{l=1}^{3}\left\{\frac{1}{h_l \cdot w_l}\sum_{i=1}^{h_l \cdot w_l} M_{loc}^l \right\}
\end{equation}

\begin{figure}[t]
  \centering
   \includegraphics[width=0.9\linewidth]{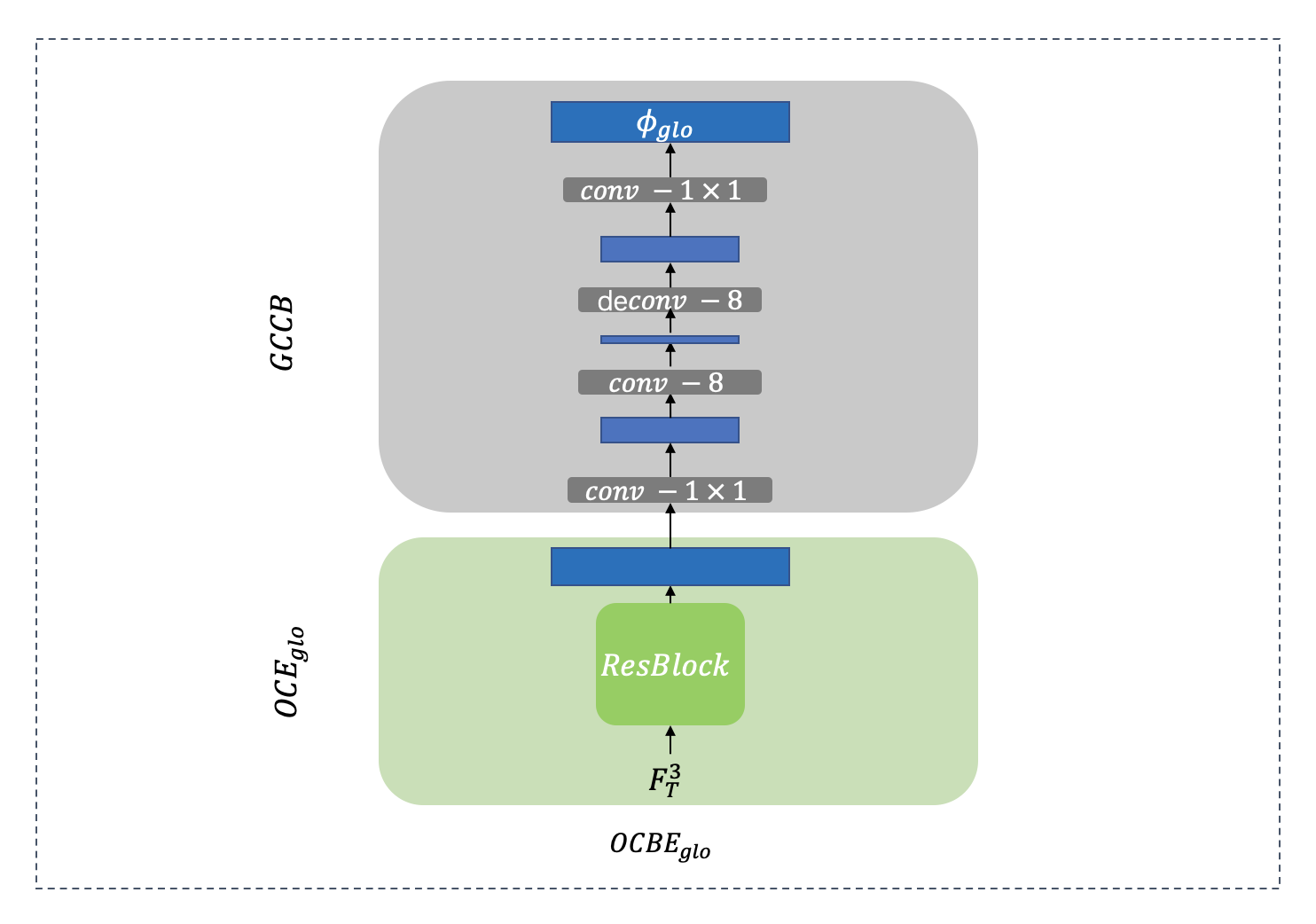}

   \caption{The one-class bottleneck embedding module for global student and the global context condensing block. We use the $4$-th residual stage of ResNet as trainable $OCE_{glo}$.}
   \label{fig_gccb}
\end{figure}

\subsection{Global Contextual Affinity Distillation}
Although the encoder-decoder architecture naturally can keep the most important information and the multi-scale feature distillation paradigm could take both low-level information and high-level information into account, the student still cannot learn globally. Furthermore, the $OCBE_{loc}$ fuses low-level features into the final embedding space, which is beneficial for accurate low-level feature reconstruction but decreases the global contextual learning ability. We design the global context condensing block to keep the most important global information as shown in Fig. \ref{fig_gccb}. It is realized by compressing the high-semantic level feature $F_T^3$ into a one-dimensional feature space with $g$ channels and then restoring it to the original feature size. The output $\phi_{glo}$ of $GCCB$ is then decoded by a student $S_{glo}$ that has the identical architecture as $S_{loc}$.

To further encourage the global student to better learn the global contextual information, different from the per-pixel cosine similarity loss used for the local student, we propose the contextual affinity loss for the global student. To learn the local feature embedding $f_{S_{loc}, i}^l$, the local student $S_{loc}$ can only learn from $f_{f_T, i}^l$, which is a benefit for accurately reconstruct local features. However, it fails to learn the contextual information from the whole image. For example, if a normal feature appears in the wrong position, the local student can't detect it as anomalous since the feature itself is a representation of a normal structure. We are inspired to propose the contextual affinity loss aiming to enable the student to learn a local feature $f_{S_{glo}, i}^l$ from the whole feature map. 
For a feature vector $f_{T, i}^l$ from a feature map extracted by the teacher $F_T^l$, we first compute the cosine similarity between $f_{T, i}^l$ and all feature vectors to get a similarity list $A_{T, i}^l =[a_{i, 1}^{t, l}, ..., a_{i, h_l \cdot w_l}^{t, l}]$, where 
\begin{equation}
    a_{i, j}^{t, l}=\frac{f_{T, i}^l \cdot f_{T, j}^l}{\|f_{T, i}^l\| \|f_{T, j}^l\|}
\end{equation}
We define the contextual affinity for a feature vector in the whole feature map as the probability distribution $P_{T, i}^l = [p_{T, i}^{1, l}, ..., p_{T, i}^{h_l \cdot w_l, l}]$ where
\begin{equation}
    p_{T, i}^{i, l} = \frac{\exp(\frac{{a_{i, j}^{t, l}}}{\mathcal{T}})}{\sum_{j=1}^{h_l \cdot w_l} \exp(\frac{{a_{i, j}^{t, l}}}{\mathcal{T}})}
\end{equation}
where $\mathcal{T}$ is the temperature. By converting the similarity list into a probability distribution, the scales of the contextual affinity for each feature vector are normalized and the large spatial similarity relations that we believe are the most important elements are paid more attention to, while the less similar relations are ignored. By using a small $\mathcal{T}$, the probability distribution becomes harder, which means we only focus on a small portion of spacial relations in the feature map. 
Similarly, for the student feature embedding, it is intuitive to compute the corresponding contextual affinity probability distribution $P_{S_{glo}, i}^l$ for $f_{S_{glo}, i}^l$ \textit{within the student feature map} $F_{S_{glo}}^l$ and minimize the difference between $P_{T, i}^l$ and $P_{S_{glo}, i}^l$. However, in this case, the optimization of one feature vector $f_{S_{glo}, i}^l$ is coupled with all the feature vectors in $F_{S_{glo}}^l$, making the optimization difficult \cite{krahenbuhl2011efficient}. For unsupervised knowledge distillation where we only have the knowledge distillation training target, we experimentally found the model can't converge. Considering that we also use the teacher model for inference, we compute the contextual similarity list $A_{S_{glo}, i}^l$ and affinity probability distribution $P_{S_{glo}, i}^l$ using the corresponding student feature embedding $f_{S_{glo}, i}^l$ and the whole \textit{teacher feature map} $F_T^l$, where
\begin{equation}
    a_{i, j}^{s_{glo}, l}=\frac{f_{S_{glo}, i}^l \cdot f_{T, j}^l}{\|f_{S_{glo}, i}^l\| \|f_{T, j}^l\|}
\end{equation}
We then use KL divergence to evaluate the discrepancy between the contextual affinity distributions from the teacher and global student
\begin{equation}
\label{kl}
    \mathcal{KL}(P_{T, i}^l, P_{S_{glo}, i}^l) =\mathcal{T}^2 \sum_{j=1}^{h_l \cdot w_l} p_{T, i}^{j, l}\cdot\log{\left[\frac{p_{T, i}^{j, l}}{p_{S_{glo}, i}^{j, l}}\right]}
\end{equation}
The $p_{T, i}^{j, l}$ in Equation \eqref{kl} could be interpreted as a weighting factor. Large $p_{T, i}^{j, l}$ values that indicate the spatial relations with high similarities are paid more attention to, while the KL divergence tends to neglect less similar relations. Note that since each feature vector always has the largest similarity with itself, it is not a contradiction with accurately reconstructing the feature vector. The high-similarity relations are the guiding \textit{signposts} for training the student feature vector from the global context. During inference, the global student fails to capture the global contextual information for logical anomalies. Similarly, the final loss for training the global student is 
\begin{equation}
    \mathcal{L}_{glo} = \sum_{l=1}^{3}\left\{\frac{1}{h_l \cdot w_l}\sum_{i=1}^{h_l \cdot w_l}\mathcal{KL}(P_{T, i}^l, P_{S_{glo}, i}^l)  \right\}
\end{equation}

\subsection{Pixel and Image Anomaly Scoring}
Following Equation \eqref{map_loc} and Equation \eqref{kl}, we could get anomaly score maps $M_{loc}^l$ and $M_{glo}^l$ for the $l$-th layer from the local and global student. Each element in the score map indicates the feature or contextual affinity discrepancy. To get precise multi-scale anomaly detection and localization, we first up-sample each score map to the image resolution and  conduct element-wise addition for each student. The final score map for an input image $I$ is the combination of the two students' normalized detection results
\begin{equation}
\begin{split}
      M(I) = \frac{M_{loc} - \mu_{loc}}{\sigma_{loc}} + \frac{M_{glo} - \mu_{glo}}{\sigma_{glo}},\\
    M_{loc} = \sum_{l = 1}^{3}\Psi(M_{loc}^{l}), M_{glo} = \sum_{l = 1}^{3}\Psi(M_{glo}^{l})  
\end{split}
\end{equation}
Where $\Psi$ is the bilinear up-sampling operation, $\mu$ and $\sigma$ are the mean and standard deviation values, respectively. They are computed on the validation set $\mathcal{S}^v$ or the training set $\mathcal{S}^t$ if $\mathcal{S}^v$ is not available. The image-level anomaly score is derived by choosing the maximum score from the final score map. We apply a Gaussian filter before image-level anomaly scoring to remove local noises.

\begin{table*}[t]
\caption{\textit{Anomaly Localization} results on MVTec LOCO AD dataset\cite{bergmann2022beyond}. The area under the sPRO curve is computed up to an average false positive rate of 0.05. We report the mean scores for structural and logical anomalies. The best scores are in bold.}
\centering
\begin{tabular}{lcccccc}
\hline
Method      & Breakfast Box  & Screw Bag      & Pushpins       & Splicing Connectors & Juice Bottle   & Mean           \\ \hline
AE          & 0.189          & 0.289          & 0.327          & 0.479               & 0.605          & 0.378          \\
VAE         & 0.165          & 0.302          & 0.311          & 0.496               & 0.636          & 0.382          \\
MNAD \cite{park2020learning}       & 0.080          & 0.344          & 0.357          & 0.442               & 0.472          & 0.339          \\
VM          & 0.168          & 0.253          & 0.254          & 0.125               & 0.325          & 0.225          \\
f-AnoGAN \cite{schlegl2019f}    & 0.223          & 0.348          & 0.336          & 0.195               & 0.569          & 0.334          \\
SPADE \cite{cohen2020sub}      & 0.372          & 0.331          & 0.234          & 0.516               & 0.804          & 0.451          \\
US \cite{bergmann2020uninformed}         & 0.496          & 0.602          & 0.523          & 0.698               & 0.811          & 0.626          \\
RD \cite{deng2022anomaly}         & 0.326          & 0.568          & 0.597          & 0.702               & 0.840          & 0.607          \\
PatchCore-25 \cite{roth2022towards} & 0.510 & 0.577 & 0.504 & 0.731 &0.794 & 0.623 \\
GCAD \cite{bergmann2022beyond}       & 0.502          & 0.558          & 0.739          & \textbf{0.798}      & \textbf{0.910} & 0.701          \\
DSKD (Ours) & \textbf{0.568} & \textbf{0.627} & \textbf{0.825} & 0.767               & 0.865          & \textbf{0.730} \\ \hline
\end{tabular}
\label{tab_al}
\end{table*}

\begin{table}[t]
\caption{The image-level \textit{anomaly detection} AUROC scores on MVTec LOCO AD dataset. The best scores are in bold.}
\centering
\resizebox{\linewidth}{!}{
\begin{tabular}{lccc}
\hline
Method   & Structural AD        & Logical AD        & Mean           \\ \hline
AE       & 0.565                & 0.581             & 0.573          \\
VAE      & 0.548                & 0.538             & 0.543          \\
MNAD     & 0.702                & 0.600             & 0.651          \\
VM       & 0.589                & 0.565             & 0.577          \\
f-AnoGAN & 0.627                & 0.658             & 0.642          \\
SPADE    & 0.668                & 0.709             & 0.689          \\
US       & \textbf{0.883}       & 0.664             & 0.773          \\
RD       & 0.867                & 0.669             & 0.768          \\
PatchCore-25  &  0.855          & 0.759      &   0.807 \\
GCAD     & 0.806                & \textbf{0.860}    & 0.833          \\
DSKD (Ours)     & 0.869                & 0.812             & \textbf{0.840} \\ \hline
\end{tabular}}

\label{tab_ad}
\end{table}

\begin{figure*}
  \centering
    \includegraphics[width=0.9\linewidth]{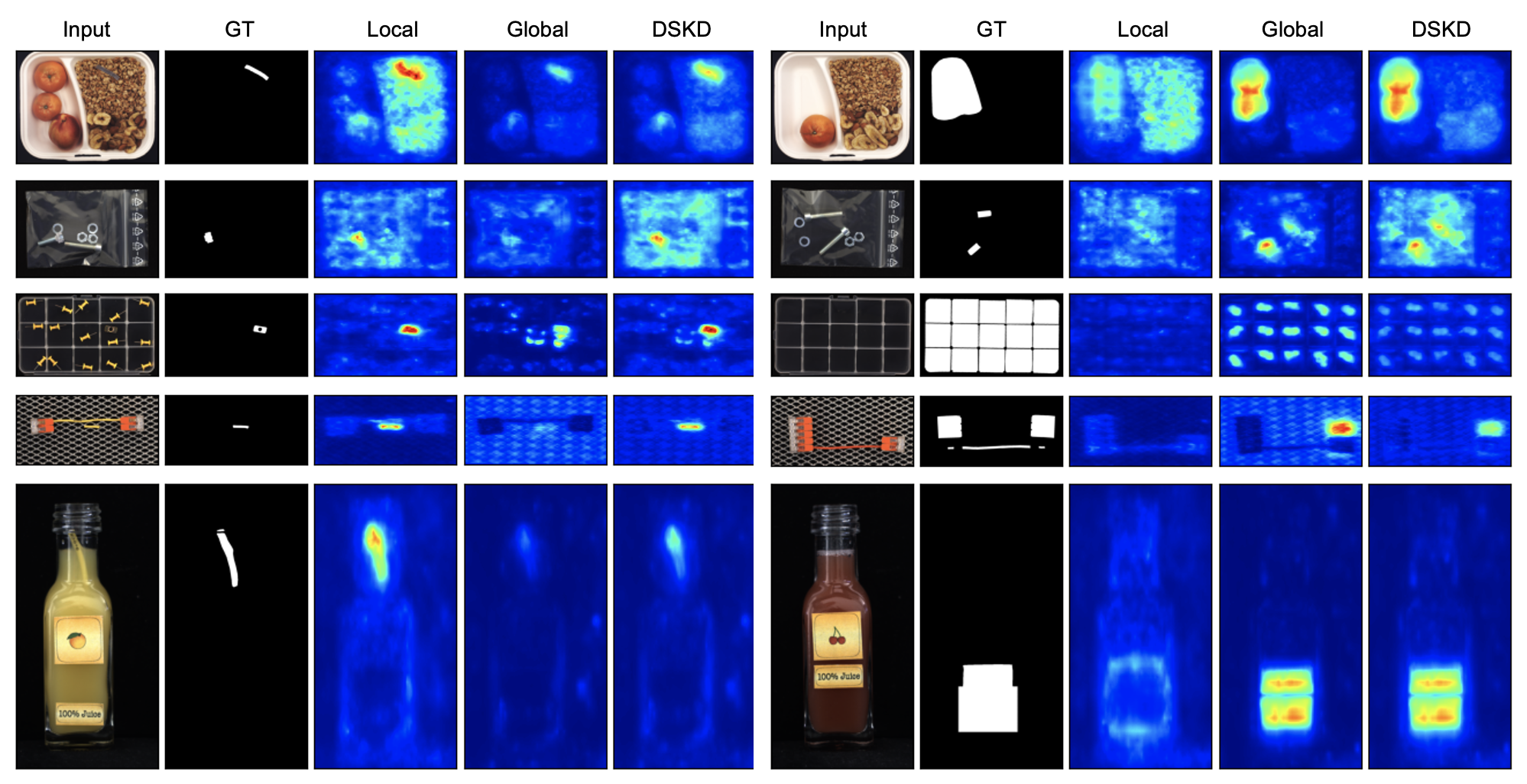}

  \caption{Qualitative examples for each category. Structural anomalies from top to bottom: "contamination" on cereals, "broken" screw head, "contamination" in one compartment, "additional" short cable, and "contamination" on the juice. Logical anomalies from top to bottom: "missing" one nectarine and one tangerine, one short screw "replaced" by a long screw, "missing" pushpins for all compartments, "wrong connector pair", and "misplaced" bottom label. We show the detection results from the local student which is identical to RD \cite{deng2022anomaly}, the global student, and the final detection results.}
    \label{fig_visual}
\end{figure*}

\section{Experimental Results}
\subsection{Experimental Settings}
\textbf{Datasets.} We use two public unsupervised anomaly detection and localization benchmarks: MVTec LOCO AD \cite{bergmann2022beyond} and the modified MVTec AD \cite{bergmann2019mvtec}. The recently introduced MVTec LOCO AD covers both local structural anomalies and logical anomalies that violate long-range dependencies. It contains 5 object categories and 1772 anomaly-free images in total for training. 304 anomaly-free images are also provided for validation. Each of the 1568 test images is either anomaly-free or contains at least one structural or logical anomaly. Pixel-level annotations are provided for testing. The MVTec AD dataset has 15 single object or texture categories, consisting mainly of anomaly-free and structural anomalous samples but 37 test images that are better defined as logical anomaly samples \cite{bergmann2022beyond} are split out as the logical anomaly test subset. \par

\textbf{Model training.} All images are resized to $256 \times 256$ resolution. We follow the one-model-per-category setting of previous studies. The two students could be trained simultaneously or separately. For each student, we use the same training configuration as \cite{deng2022anomaly}. We use Adam optimizer using $\beta=(0.5,0.999)$ with a fixed learning rate $0.005$. Each student is trained for $200$ epochs with the same batch size of $16$. The channel dimension $g$ of $GCCB$ is set to 1024 by default and the temperature $\mathcal{T}$ is set to 1.\par

\textbf{Evaluation metrics.} We take the area under the receiver operating characteristic (AUROC) score as the threshold-free image-level anomaly detection evaluation metric. For anomaly localization, it is also suitable to use AUROC for structural anomaly detection evaluation. However, logical anomalies, {\textit{e}.\textit{g}.} a missing object, are difficult to annotate and segment for each pixel. We use the saturated per-region overlap (sPRO)\cite{bergmann2022beyond}, a generalized version of the PRO metric \cite{bergmann2019mvtec} to evaluate the anomaly localization performance. The metric score saturates once the overlap with the ground truth meets a predefined saturation threshold. All thresholds are also provided by LOCO dataset.

\subsection{Experimental Results on LOCO}
We compare our proposed method against autoencoders including a vanilla autoencoder (AE), a variational autoencoder (VAE), a memory-guided autoencoder (MNAD) \cite{park2020learning}, f-AnoGAN \cite{schlegl2019f}, Variation Model (VM)\cite{steger2018machine}, Uninformed Students (US) \cite{bergmann2020uninformed}, SPADE \cite{cohen2020sub}, Reverse Distillation (RD) \cite{deng2022anomaly} and Global Context Anomaly Detection (GCAD) \cite{bergmann2022beyond}. The same data augmentations are used as GCAD \cite{bergmann2022beyond} throughout our experiments. 

We first report the anomaly localization results in Table \ref{tab_al}. The average score of our proposed method over 5 categories reaches 0.73 at a very low integration limit of 0.05. Especially on the breakfast box, screw bag, and pushpins that most of our counterparts can only reach relatively low scores because of the complex contextual logical constraints, we outperformed their performance by a large margin.

The image-level anomaly detection results are shown in Tab. \ref{tab_ad}. Most of the existing methods could already achieve relatively high performance on structural anomaly detection, especially the knowledge distillation-based US \cite{bergmann2020uninformed} and RD \cite{deng2022anomaly} that use patch-based and per-pixel training targets. However, the logical anomaly detection scores drop fast compared to their good performance on structural anomaly detection. Although GCAD \cite{bergmann2022beyond} is based on US \cite{bergmann2020uninformed} aiming to improve the logical anomaly detection ability and achieves the best score on it, the performance on structural anomaly detection is severely cumbered. The proposed method improves the logical anomaly detection performance by a large margin, without decreasing RD \cite{deng2022anomaly} 's structural anomaly detection ability and also achieves a new state-of-the-art (0.84 score).

We give some qualitative visualization results for each category in Fig. \ref{fig_visual}. For each category, we give one structural anomaly image on the left and one logical anomaly image on the right. The local student is mainly responsible for low-semantic level structural anomaly detection but fails to capture long-range dependencies, while the global student could better learn the global contextual constraints but can not perform well on fine-grained local structural anomaly detection. Finally, the dual-student knowledge distillation framework enables our method to detect both structural anomalies and logical anomalies.

\begin{table}[t]
\caption{Experimental results on the modified MVTec AD \cite{bergmann2019mvtec}. We report the image-level AUROC scores / the normalized AU sPRO scores with an integration limit of 0.05. The best results are in bold and the second-best results are with underlines.}
\centering
\resizebox{\linewidth}{!}{
\begin{tabular}{lccc}
\hline
Method   & Structural   & Logical     & Mean                   \\ \hline
AE       & 0.761/0.337          & 0.718/0.224          & 0.740/0.281          \\
VAE      & 0.766/0.336          & 0.737/0.215          & 0.751/0.276          \\
MNAD     & 0.709/0.294          & 0.427/0.032          & 0.568/0.163          \\
VM       & 0.690/0.240          & 0.679/0.069          & 0.684/0.155          \\
f-AnoGAN & 0.751/0.290          & 0.751 0.231          & 0.751/0.261          \\
SPADE    & 0.898/0.632          & 0.906/0.647          & 0.902/0.640          \\
US       & 0.936/0.762          & 0.747/0.417          & 0.842/0.590           \\
RD       & \textbf{0.986}/\textbf{0.793}          & 0.914/0.477          & \underline{0.950}/0.635 \\
PatchCore-25 & \underline{0.985}/\underline{0.790}      & \textbf{0.998}/0.619          & \textbf{0.992}/\underline{0.705} \\
GCAD     & 0.871/0.716          & \underline{0.991}/\textbf{0.863}          & 0.931/\textbf{0.789}  \\
DSKD (Ours)    & 0.955/0.755    & 0.906/\underline{0.649}          & 0.931/0.702         \\ \hline
\end{tabular}}

\label{tab_mvtecad}
\end{table}

\subsection{Experimental Results on modified MVTec AD}
We report the anomaly detection and localization results on the modified MVTec AD dataset in Tab. \ref{tab_mvtecad}. The results show that some of the existing works perform well on structural anomaly detection, while still showing the ability for logical anomaly detection. The SPADE \cite{cohen2020sub}achieves a 0.906 image-level AUROC score which is even higher than structural anomaly detection. An underlying assumption is that SPADE uses the high-semantic level feature, \textit{e}.\textit{g}., the output of the 4-th residual block of a pre-trained ResNet for image-level anomaly scoring. RD \cite{deng2022anomaly} also achieves a high score of 0.914 because it benefits from the compact one-class bottleneck embedding space that also contains high-semantic level information. PatchCore achieves SOTA image-level anomaly detection results with the help of using locally aware patch features. Similar to the results on LOCO, GCAD \cite{bergmann2022beyond} is capable of logical anomaly detection, at the cost of an obvious performance drop for structural anomaly detection. The proposed DSKD showed better logical and average anomaly localization performance. 

\begin{figure}
  \centering
    \includegraphics[width=0.9\linewidth]{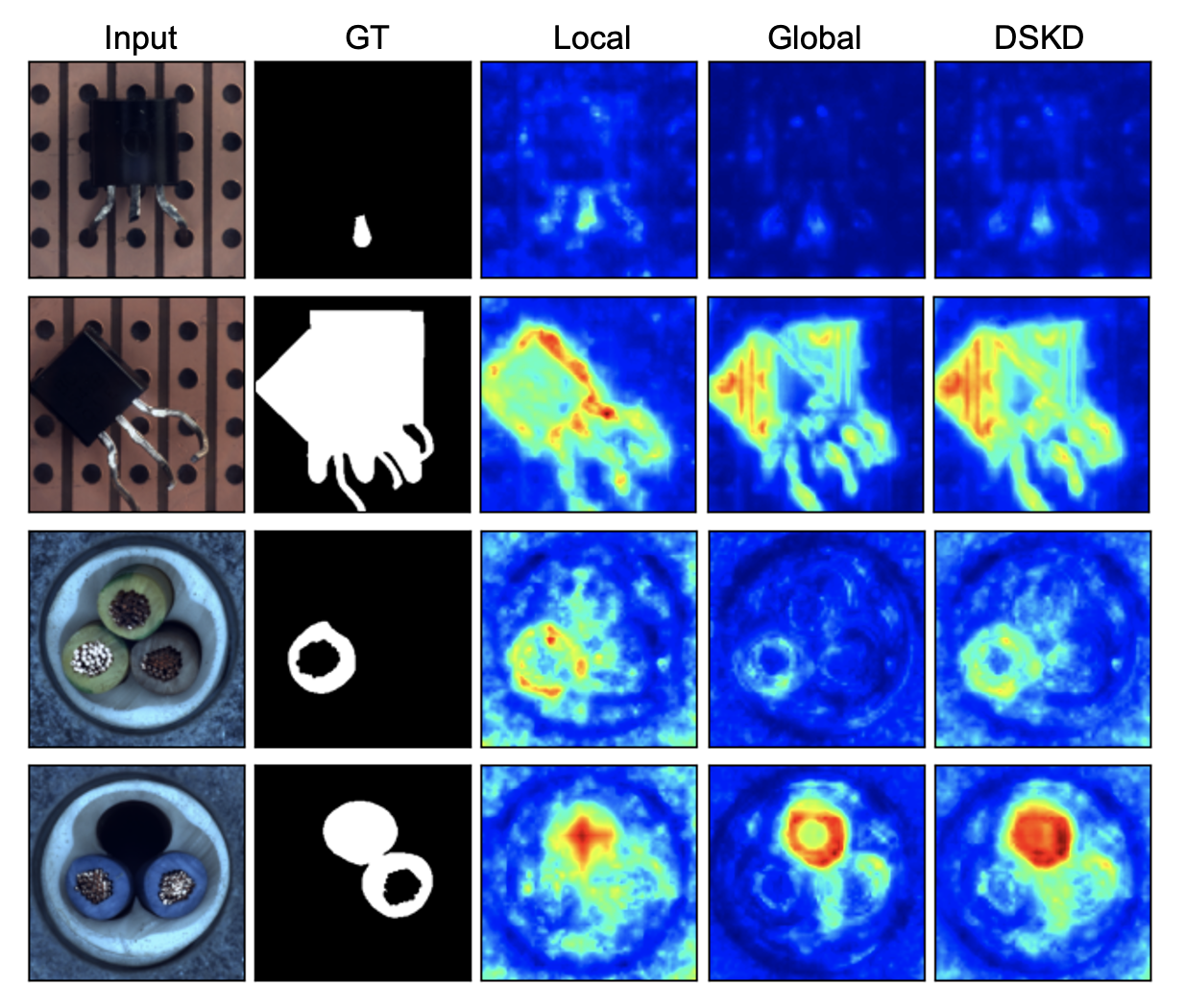}

  \caption{Qualitative results of logical anomaly detection on the modified MVTec AD dataset.}
    \label{fig_mvtec}
\end{figure}

We visualize two logical anomaly samples from the transistor and cable category in Fig \ref{fig_mvtec}, where one example is better detected by the local student while the remaining one is better detected by the global student.  Although our proposed method achieves the second-best overall performance by improving the logical anomaly detection ability without obviously affecting the structural anomaly detection performance, we observed similar limitations with results on the LOCO dataset. Our global student could capture global logical constraints but is not sensitive to small-sized defects and ambiguous anomalies violating both low-level and long-range dependencies. In Fig. \ref{fig_mvtec}, a blue cable replaced by a green one may also be defined as a kind of color contamination. But the global student could identify a missing object or an object in the wrong place while identifying the right position.

\subsection{Ablation Studies}

We investigate the effectiveness of the dual-student architecture, and the contextual affinity loss and assess the sensitivity of hyperparameters. The performance of a single student trained with different losses is reported in Tab. \ref{single_student}. Benefiting from the RD \cite{deng2022anomaly} architecture which has a low-level feature bias, and our contextual affinity learning scheme, the local student is capable of logical anomaly detection and yields the best overall performance. The global student trained with per-pixel cosine similarity is enhanced for better low-level feature reconstruction which in turn improved both structural and logical anomaly detection performance.
\begin{table}[t]
\caption{Anomaly detection results using different student architecture and loss. Pi means per-pixel cosine similarity loss and CA means our proposed contextual affinity loss.}
\label{single_student}
\centering
\begin{tabular}{lcccc}
\hline
Model  & Loss Type & Structural & Logical & Mean           \\ \hline
Local  & Pi        & 0.739              & 0.474           & 0.607          \\
Local  & CA        & \textbf{0.752}     & 0.507           & \textbf{0.630} \\
Global & Pi        & 0.547              & \textbf{0.693}  & 0.620          \\
Global & CA        & 0.336              & 0.640           & 0.488          \\ \hline
\end{tabular}

\end{table}

Tab. \ref{loss_pair} gives qualitative comparisons of our dual-student architecture trained with different loss pairs. Although the global student trained with per-pixel loss outperforms the one trained with contextual affinity loss, however, the dual-student architecture design releases the constraint of accurate low-level feature reconstruction for the global student and encourages the global student to focus on global contextual information.
\begin{table}[t]
\caption{Anomaly detection results with different loss combinations for the dual-student knowledge distillation framework. }
\centering
\begin{tabular}{lcccc}
\hline
$\mathcal{L}_{loc}$ & $\mathcal{L}_{glo}$ & Structural & Logical & Mean           \\ \hline
Pi    & Pi    & 0.748              & 0.675           & 0.711          \\
CA    & CA    & 0.752              & 0.678           & 0.708          \\
Pi    & CA    & \textbf{0.754}     & \textbf{0.707}  & \textbf{0.730} \\ \hline
\end{tabular}

\label{loss_pair}
\end{table}

We also investigate the impact of $GCCB$ along with its channel dimensions $g$. The results are shown in Tab. \ref{g_value}. The use of $GCCB$ improved the performance by a large margin and performs well with various $g$ values.
\begin{table}[t]
\caption{Mean detection results with different $g$ dimension values. "w/o" means $GCCB$ is not used and 
 for $g = 2048$ channels, we do not use $conv 1 \times 1$ layers to downsample and upsample the channel dimensions.}
\centering
\resizebox{\linewidth}{!}{
\begin{tabular}{l|cccccc}
\hline
$g$       & w/o & 512   & 768   & 1024           & 1280  & 2048  \\ \hline
AU sPro & 0.607            & 0.714 & 0.721 & \textbf{0.730} & 0.720 & 0.729 \\ \hline
\end{tabular}}

\label{g_value}
\end{table}

\begin{figure}
  \centering
    \includegraphics[width=0.8\linewidth]{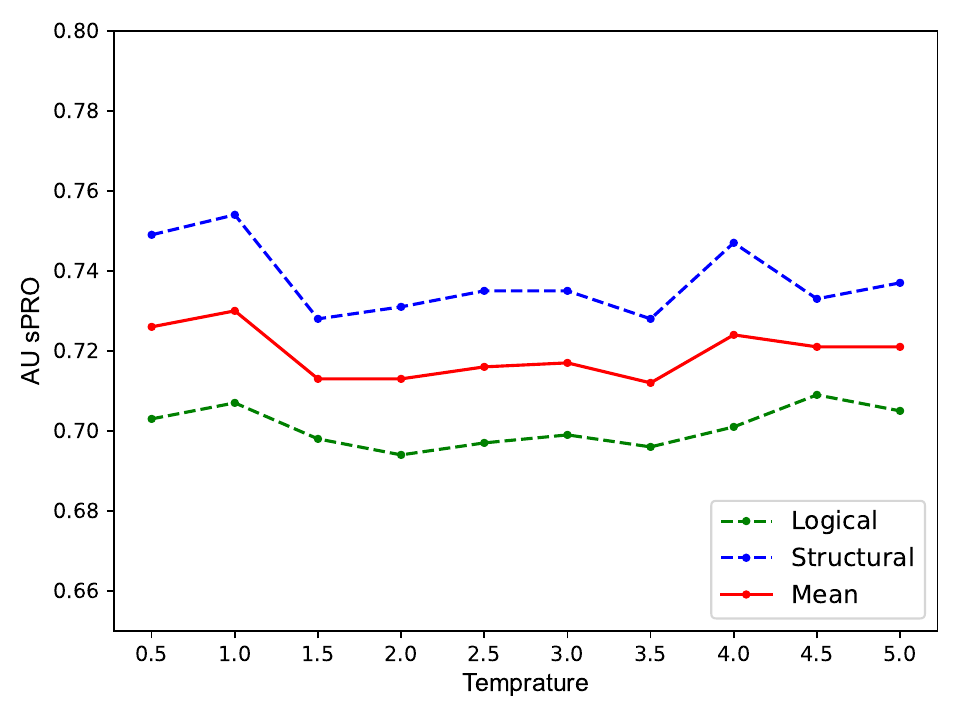}
  \caption{Impact of temperature $\mathcal{T}$.}
    \label{fig_t}
\end{figure}

The results with different $\mathcal{T}$ are shown in Fig. \ref{fig_t}. A large $\mathcal{T}$ makes the distribution softer and covers wider relations. Although it may confront the low-level feature reconstruction ability, our method is stable for a wide range of $\mathcal{T}$.

\section{Conclusion}
We proposed the dual-student knowledge distillation framework and contextual affinity loss for anomaly detection. The two students play different roles and the contextual affinity computation for the student model uses both teacher and student features. The design and use of global context condensing block $GCCB$ and the contextual affinity loss enable our method capable of both structural and logical anomaly detection. Experiments showed that the proposed method outperformed previous studies and achieved SOTA performance on public benchmarks.

{\small
\bibliographystyle{ieee_fullname}
\bibliography{egbib}
}

\end{document}